\documentclass[journal]{IEEEtran}

\usepackage{cite}
\usepackage[pdftex]{graphicx}
\usepackage{comment}
\usepackage{amsmath,amssymb} 
\usepackage{color}
\usepackage{epsfig}
\usepackage{algorithm}
\usepackage{algorithmic}
\usepackage{multirow}
\usepackage{booktabs}
\usepackage{float}
\usepackage{caption}
\usepackage{latexsym}
\usepackage{array}
\newcommand{\PreserveBackslash}[1]{\let\temp=\\#1\let\\=\temp}
\newcolumntype{C}[1]{>{\PreserveBackslash\centering}p{#1}}
\newcolumntype{R}[1]{>{\PreserveBackslash\raggedleft}p{#1}}
\newcolumntype{L}[1]{>{\PreserveBackslash\raggedright}p{#1}}

\author{Chen Ju,
        Peisen~Zhao,
        Siheng~Chen, Member, IEEE,
        Ya Zhang, Member, IEEE,\\
        Xiaoyun~Zhang,
        and Qi Tian, Fellow, IEEE
\IEEEcompsocitemizethanks{\IEEEcompsocthanksitem 
This work is supported by the National Key Research and Development Program of China (No. 2020YFB1406801), 111 plan (No. BP0719010),  and STCSM (No. 18DZ2270700), and State Key Laboratory of UHD Video and Audio Production and Presentation. (Corresponding author: Ya Zhang)

C. Ju, P. Zhao, S. Chen, Y. Zhang and X. Zhang are with the Cooperative Medianet Innovation Center, Shanghai Jiao Tong University, Shanghai 200240, China. (E-mail: \{ju\_chen, pszhao, sihengc, ya\_zhang, xiaoyun.zhang\}@sjtu.edu.cn).

Q. Tian is with the Huawei Noah’s Ark Lab, Shenzhen, Guangdong 518129, China. (E-mail: tianqi1@huawei.com).}}

\begin{document}

\title{Adaptive Mutual Supervision for\\Weakly-Supervised Temporal Action Localization}

\markboth{Journal of \LaTeX\ Class Files,~Vol.~14, No.~8, August~2015} 
{Shell \MakeLowercase{\textit{et al.}}: Bare Demo of IEEEtran.cls for IEEE Journals}

\maketitle

\begin{abstract}
Weakly-supervised temporal action localization aims to localize actions in untrimmed videos with only video-level action category labels. Most of previous methods ignore the incompleteness issue of Class Activation Sequences (CAS), suffering from trivial localization results. To solve this issue, we introduce an adaptive mutual supervision framework (AMS) with two branches, where the base branch adopts CAS to localize the most discriminative action regions, while the supplementary branch localizes the less discriminative action regions through a novel adaptive sampler. The adaptive sampler dynamically updates the input of the supplementary branch with a sampling weight sequence negatively correlated with the CAS from the base branch, thereby prompting the supplementary branch to localize the action regions underestimated by the base branch. To promote mutual enhancement between these two branches, we construct mutual location supervision. Each branch leverages location pseudo-labels generated from the other branch as localization supervision. By alternately optimizing the two branches in multiple iterations, we progressively complete action regions. Extensive experiments on THUMOS14 and ActivityNet1.2 demonstrate that the proposed AMS method significantly outperforms the state-of-the-art methods.
\end{abstract}


\IEEEpeerreviewmaketitle

\section{Introduction}
\label{section:introduction}
\IEEEPARstart{T}{emporal} action localization, which localizes actions from untrimmed videos, plays an important role in video understanding. Although several studies~\cite{chao2018rethinking,lin2018bsn,lin2019bmn,shou2016temporal,gao2017turn,lin2019fast,zhao2017temporal} have shown promising results on strongly-supervised temporal action localization, the annotations in the form of precise action boundaries are both time-consuming and noisy. Weakly-supervised temporal action localization (WTAL), which handles the same problem but only requires video-level action category labels, has recently received increasing attention~\cite{nguyen2018weakly,lee2019background,paul2018w,shi2020weakly,shou2018autoloc,zeng2019breaking,su2018cascaded,liu2019completeness}.

\begin{figure}[t]
\begin{center}
\includegraphics [width=0.49\textwidth] {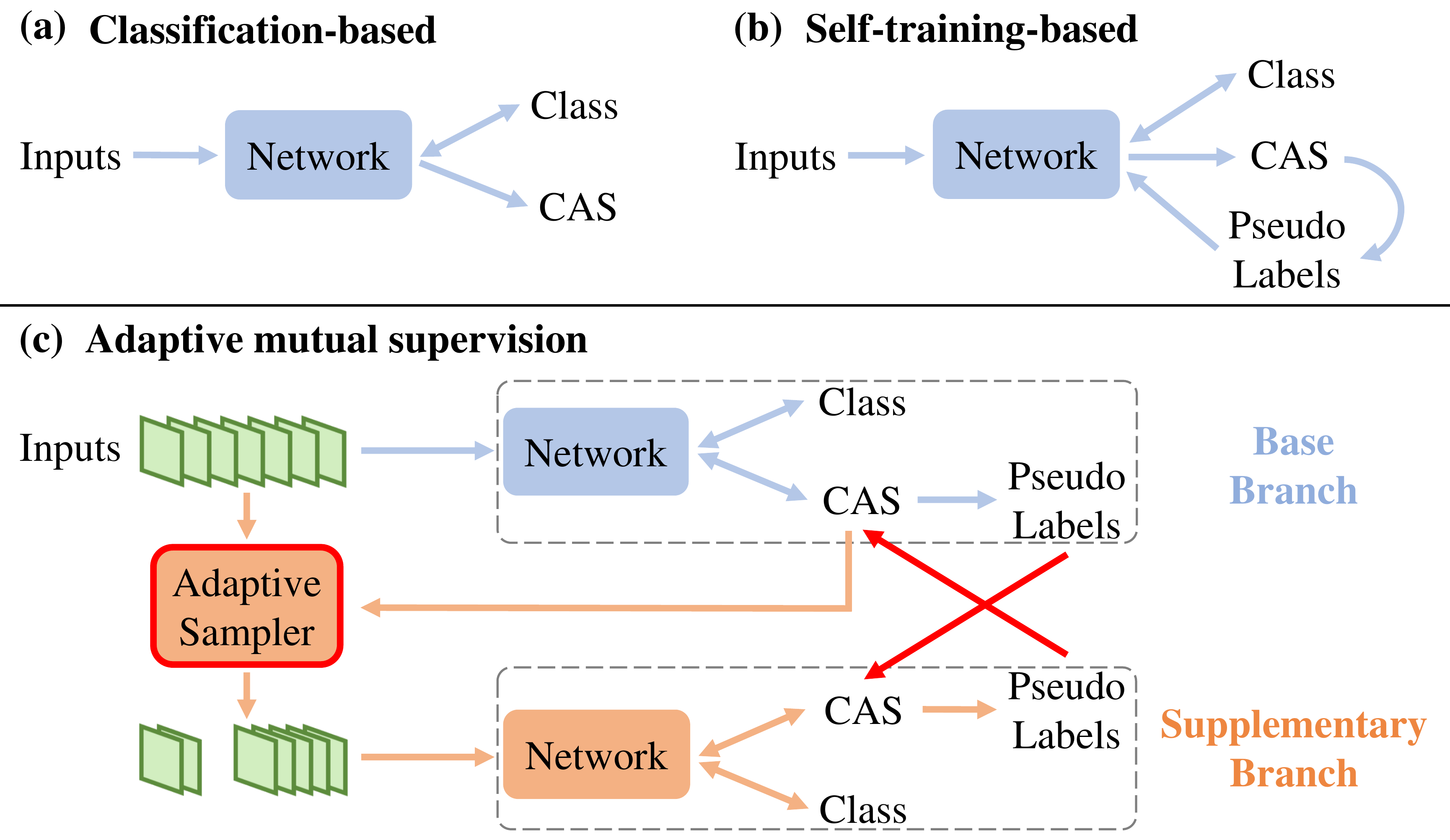}
\end{center}
\vspace{-2pt}
\caption{Framework comparison in WTAL. `Class' is classification supervision, `CAS' is class activation sequence, and arrows denote the propagation direction.
\textbf{(a): Classification-based framework}~\cite{nguyen2018weakly,lee2019background,paul2018w} thresholds CAS for localization. \textbf{(b): Self-training-based framework}~\cite{pardo2021refineloc,zhai2020two,luo2020weakly} relies on location pseudo-labels generated from CAS. Both of these frameworks ignore the incompleteness issue of CAS, suffering from trivial localization results. \textbf{(c): Adaptive mutual supervision framework}. Red is our two key contributions.}
\label{fig:introduce}
\end{figure}

To date in the studies, there are two main frameworks in the WTAL task. The classification-based framework~\cite{nguyen2018weakly,shou2018autoloc,lee2019background,paul2018w} adopts the idea of multiple instance learning (MIL), \emph{i.e.}, first training an action classifier by video-level category labels, then thresholding the class activation sequence (CAS) of the classifier to obtain action proposals; see Fig.\,\ref{fig:introduce}\,(a). This framework only optimizes the classification objective. On the other hand, regarding the CAS as a noisy location cue, the self-training-based framework~\cite{pardo2021refineloc,zhai2020two,luo2020weakly} iteratively thresholds the CAS of the current step to generate location pseudo-labels for the next step, and progressively refines the action localization results; see Fig.\,\ref{fig:introduce}\,(b).

The CAS generated from the classifier, indicating the class-specific action probability of each snippet, becomes the key to the localization performance of the above two frameworks. However, CAS has an \textbf{incompleteness issue}, \emph{i.e.}, it only covers the most discriminative regions that contribute most to action classification~\cite{zeng2019breaking,su2018cascaded,liu2019completeness}. Since there is a fundamental difference in optimization objectives between classification and localization, \emph{i.e.}, classification mainly relies on the most discriminative action regions while localization requires mining complete action regions, CAS is usually sparse and incomplete. As a result, the action proposals and the location pseudo-labels produced from CAS are both low-quality, causing trivial localization results in these two frameworks.

To solve the incompleteness issue, this paper considers a novel adaptive mutual supervision framework (AMS) with two branches that are collaborative and complementary. The base branch adopts the CAS to localize the most discriminative action regions, which is similar to above two frameworks. While the supplementary branch localizes the less discriminative action regions to complete localization results of the whole framework. To achieve collaboration and complementarity, we propose two core designs: an \textbf{adaptive sampler} and \textbf{mutual location supervision}; see Fig.\,\ref{fig:introduce}\,(c).

The design rationale of the adaptive sampler is to select the less discriminative regions underestimated by the base branch as inputs for the supplementary branch, so that the supplementary branch can focus on these challenging action regions. Concretely, we feed the original video into the base branch, while leverage the adaptive sampler to probabilistically select video snippets for the supplementary branch. Since the CAS of the base branch mainly localizes the most discriminative action regions, the sampling probability sequence is designed to be dynamic and negatively correlated with the CAS from the base branch. That is, we over-sample the snippets with low CAS values while under-sample those with high CAS values. As a result, the inputs of the supplementary branch mainly consist of the snippets corresponding to the low CAS regions of the base branch, which prompts the supplementary branch to purposefully mine less discriminative action regions, and thus completes detected action information.

To further promote mutual enhancement between the base branch and the supplementary branch, we are motivated to design mutual location supervision, which forces each branch to explicitly optimize the localization objective with location pseudo-labels from the other branch. Specifically, both the two branches use video-level category labels as classification supervision; meanwhile, each branch leverages location pseudo-labels generated from the CAS of the other branch as localization supervision. For optimization, we alternately freeze one branch and train the other branch. In this process, the localization results of each branch are taken as the localization objective of the other branch, so that the complementary action regions of the two branches are combined to make the localization supervision more complete and precise.

To optimize the whole framework, we apply multiple iterations, since one single iteration brings limited improvement to excavate less discriminative action regions. In each iteration, the adaptive sampler differentiates the inputs of the two branches, so that they purposefully focus on different action regions. Then, mutual location supervision obtains more complete location supervision by pushing the CASs of the two branches to be consistent. In the next iteration, the consistent CASs in turn force the adaptive sampler to further update the inputs of the supplementary branch, thereby exploring more missing action regions. Consequently, the adaptive sampler and mutual location supervision jointly contribute to more complete results in the progressive iterations.

\textcolor{black}{In summary, our contributions are as follows. 
\begin{itemize}
  \item [1)] 
   We introduce an adaptive mutual supervision framework (AMS) for weakly-supervised temporal action localization. AMS contains a base branch and a supplementary branch, both of which generate location pseudo-labels for progressive iterative refinement.
  \item [2)]
   We design a novel adaptive sampler, which encourages the supplementary branch to further detect the underestimated action regions of the base branch, thus making the localization results more complete.
  \item [3)]
   We propose a novel mutual location supervision, which forces each branch to use location pseudo-labels obtained from the other branch, promoting mutual enhancement.
  \item [4)]
   We verify the effectiveness of AMS on two widely used benchmarks, THUMOS14~\cite{jiang2014thumos} and ActivityNet1.2~\cite{caba2015activitynet}. Our AMS method outperforms previous state-of-the-art methods, both quantitatively and qualitatively.
\end{itemize}}

We organize the rest of this paper as follows. For a better understanding of our motivation on adaptive mutual supervision, we review the related work in Section~\ref{section:relatedwork}. In Section~\ref{section:method}, we propose the novel strategies about adaptive sampler and mutual location supervision. Section~\ref{section:experiments} validates the proposed method by comparing it with existing methods. We further perform extensive ablation studies to reveal the effectiveness of each component. Finally, the conclusion is drawn in Section~\ref{section:conclusion}.

\section{Related Work}  \label{section:relatedwork}
This section reviews previous works that motivate the proposed method. We can divide those works into three groups: strongly-supervised temporal action localization, weakly-supervised temporal action localization, and adaptive sampling strategy. We next review the previous works respectively.

\subsection{Strongly-Supervised Temporal Action Localization}
Strongly-supervised temporal action localization relies on precise boundary labels and action category labels to localize action instances. The popular solutions can be summarized as the top-down framework and the bottom-up framework. The top-down framework~\cite{shou2016temporal,shou2017cdc,lin2017single,xu2017r,gao2017turn,long2019gaussian,chao2018rethinking,song2018temporal,guo2018fully,sun2021exploiting,zhou2020temporal,liu2019multiscale} first pre-defines massive anchors according to the prior knowledge of action distribution; then adopts fixed-length sliding windows to generate initial proposals; finally utilizes a boundary adjustment module to refine results. Typically, TURN~\cite{gao2017turn} and TAL-Net~\cite{chao2018rethinking} explored the effect of contextual information and the dilated temporal convolution on localization performance, respectively. The bottom-up framework~\cite{zhao2017temporal,lin2018bsn,lin2019bmn,lin2019fast,zhao2020bottom,bai2020boundary,xu2020g,gao2020accurate,long2019coarse} first predicts actionness or boundary probabilities for all video snippets, then groups the obtained start points and end points to produce proposals. Typically, BMN~\cite{lin2019bmn} listed all possible proposal groups, and ranked each proposal by evaluating the IoU between the proposal and the ground truth. BUMR~\cite{zhao2020bottom} proposed to construct constraints between the action, start, and end curves to reduce invalid proposal groups. BC-GNN~\cite{bai2020boundary} introduced graph convolution operations~\cite{schlichtkrull2018modeling} to group the most suitable start and end points. G-TAD~\cite{xu2020g} modeled each video as a graph based on temporal and semantic relationships to enhance the continuity between snippet features.
In general, the top-down framework completely discovers most action instances with few omissions, while the bottom-up framework flexibly adjusts the boundary and produces more precise predictions. For better performance, CTAP~\cite{gao2018ctap}, MGG~\cite{liu2019multi}, AFNet~\cite{chen2020afnet}, and A2Net~\cite{yang2020revisiting} further designed four fusion methods to combine these two frameworks in a supplementary manner. PBRNet~\cite{liu2020progressive} proposed a coarse-to-fine strategy to progressively refine boundaries in an end-to-end fashion. To better model the relationship between action proposals, P-GCN~\cite{zeng2019graph} and RAM~\cite{chen2019relation} utilized the graph convolution and self-attention mechanism to construct non-local networks for better feature embedding, respectively.

However, all these strongly-supervised methods demand precise action location annotations, which is time-consuming and less practical in real-world scenarios. To reduce the annotation cost, this work aims to study the same problem in the weakly-supervised setting.

\subsection{Weakly-Supervised Temporal Action Localization}
Weakly-supervised temporal action localization (WTAL) only requires video-level action category labels for training. Inspired by Class Activation Map~\cite{zhou2016learning} in object detection, early methods~\cite{wang2017untrimmednets,paul2018w,nguyen2018weakly} usually adopted the classification-based framework. That is, use the video-level category labels to train an action classifier; calculate Class Activation Sequence (CAS) based on the parameters of the classifier; post-process CAS for final action proposals.

Due to the objective difference between classification and localization, the CAS generated from the classifier has a serious incompleteness issue: it only contains local and sparse action regions. To solve this issue, SSE~\cite{zhong2018step}, CPMN~\cite{su2018cascaded}, WO~\cite{zeng2019breaking}, and A2CL-PT~\cite{min2020adversarial} introduced the erase strategies, which cascade multiple classifiers, erase the most discriminative regions detected by the previous classifier in turn, and input the remaining regions for the latter classifier, thus gradually detecting less discriminative regions. CMCS~\cite{liu2019completeness} trained multiple classifiers in parallel, which are used to detect different action regions. Since the CAS also has some false-positive activation and action-context confusion, BM~\cite{nguyen2019weakly} and BaSNet~\cite{lee2019background} proposed two background modeling methods. DGAM~\cite{shi2020weakly} explored separating context and action via Conditional Variational Auto-Encoder. In terms of post-processing CAS, Autoloc~\cite{shou2018autoloc} and KT-MGFN~\cite{su2020transferable} designed the outer-inner contrastive loss to replace the simple threshold operation. CleanNet~\cite{liu2019weakly} further proposed the action proposal evaluator for an effective boundary adjustment.

Considering that all the above methods localize with only classification supervision, some recent studies~\cite{pardo2021refineloc,luo2020weakly,zhai2020two} introduced the self-training-based framework, to provide explicit localization supervision for better localization. Its common practice is to empirically set thresholds on the CAS of the current step, and generate pseudo-labels as the location supervision of the next step; then perform several iterations to directly optimize the localization objective and progressively refine the pseudo-labels. Specifically, Refineloc~\cite{pardo2021refineloc} was the first to introduce location pseudo-labels for the WTAL task, and explored various pseudo-label generation strategies. EM-MIL~\cite{luo2020weakly} utilized class-specific CAS and class-agnostic attention as pseudo-labels, then formulated the WTAL task as an expectation-maximization problem for optimization. TSCN~\cite{zhai2020two} predicted pseudo-labels based on RGB and flow data respectively, then late fused these two pseudo-labels to alleviate false-positive results. Overall, our AMS framework differs from these studies from the following two aspects: (i) We design an adaptive sampler to encourage the supplementary branch to further detect the underestimated action regions of the base branch, which effectively completes the localization results of the whole framework; while the previous studies ignore the incompleteness of CAS, and only rely on low-quality pseudo-labels for iterative refinement; (ii) we construct mutual location supervision between two branches; while the previous studies use a single branch, whose generated location pseudo-labels provide self supervision.

\begin{figure*}[t]
\begin{center}  
\includegraphics [width=0.95\textwidth] {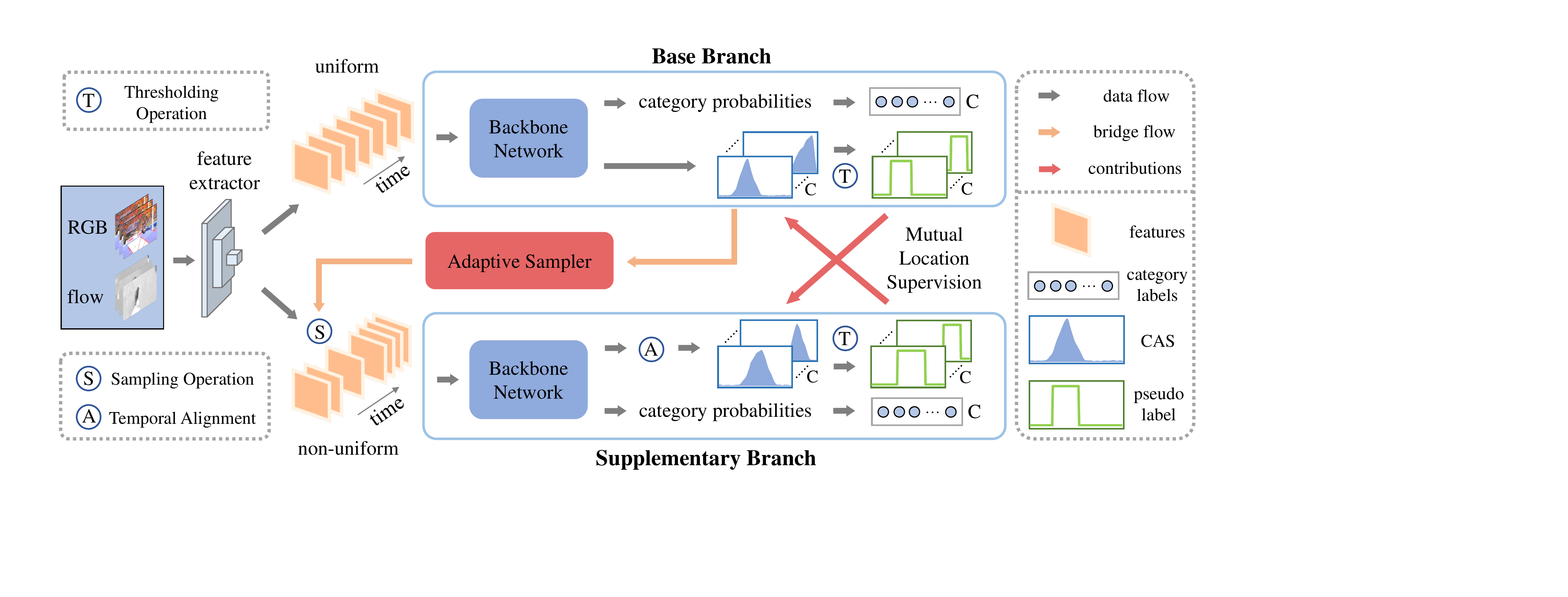}
\end{center}
\vspace{-5pt}
\caption{Framework pipeline. The pre-trained feature extractor extracts the original video features, and the backbone network predicts video category probabilities and CAS. The base branch is fed with the original video features to localize the most discriminative action regions. Then, based on the CAS of the base branch, the adaptive sampler selects the less discriminative snippet features as the inputs for the supplementary branch, which forces the supplementary branch to further detect the underestimated action regions of the base branch. Finally, each branch generates location pseudo-labels from CAS, and provides mutual location supervision during multiple training iterations.}
\label{fig:framework}
\end{figure*}
\vspace{5pt}

\subsection{Adaptive Sampling Strategy}
The role of the adaptive sampling strategy is to enlarge the local area of the image or video, thus forcing the model to focus more on some specific details. It has wide applications in fine-grained recognition, image retargeting, and small object detection. Concretely, in fine-grained image recognition, SSampler~\cite{recasens2018learning} proposed to sample based on saliency maps for data augmentation. S3N~\cite{ding2019selective} leveraged class response maps as guidelines for sampling, and achieved considerable improvements. For faster action recognition, SCSampler~\cite{korbar2019scsampler} selected a small subset of salient snippets to replace the entire video through a lightweight sampler. For better video representation, based on the importance of video frames, Coarse-Fine~\cite{kahatapitiya2021coarse} performs dynamic sampling to form different abstractions of time resolution. In image retargeting, EBID~\cite{karni2009energy} and NCV~\cite{wolf2007non} used the adaptive sampling strategy to formulate the task as the energy minimization and finite element problem. Inspired by the above studies, this paper adopts a novel adaptive sampler to differentiate the inputs of the two branches, so that they can localize different action regions. To the best of our knowledge, this is the first attempt to introduce the adaptive sampling strategy to the WTAL task.

\section{Adaptive Mutual Supervision} \label{section:method}
In this section, we propose the adaptive mutual supervision framework (AMS); see the framework pipeline in Fig.~\ref{fig:framework}. We first the formulate weakly-supervised temporal action localization problem; then, present the adaptive sampler and the mutual location supervision strategy; and finally, we introduce the training and testing details.

\subsection{Problem Formulation}
Suppose that we are given $N$ untrimmed videos $\{{v_i}\}_{i=1}^N$ and their corresponding video-level category labels $\{\mathbf{y}_i\}_{i=1}^N$, where $\mathbf{y}_i$ is a $C$-dimensional binary vector ($C$ is the total number of action categories), with $y_{i}^{k}$=1 if the $i$-th video contains the $k$-th action category, and $y_{i}^{k}$=0 otherwise. Note that each video may contain multiple action categories and multiple action instances. Our goal is to predict the temporal locations of these action categories in the video, in terms of a set of quadruples $\{(s,e,c,p)\}$, where $s$, $e$, $c$, $p$ represent the start time, the end time, the action category and the localization score of the action proposal, respectively.

Following recent methods~\cite{paul2018w,liu2019completeness,lee2019background,shi2020weakly}, for each video, we sample $T$ consecutive snippets to make sure all videos have the same length. Then, we adopt a pre-trained feature extractor to obtain the original feature $\mathbf{F^\mathrm{orig}} \in\mathbb{R}^{T\times{D}}$ for each video, where $D$ is the feature dimension of each snippet.

\subsection{Overall Framework}
The proposed AMS framework contains the base branch and the supplementary branch with identical backbones. Each backbone predicts the video category probability and respective CAS. The base branch is fed the original video feature to localize the most discriminative action regions from CAS. To strengthen collaboration and complementarity between the two branches so that the whole framework contains more complete action information, we encourage the supplementary branch to localizes the less discriminative action regions through a novel \textbf{adaptive sampler}. The sampler adaptively updates the inputs of the supplementary branch with a sampling weight sequence negatively correlated with the CAS of the base branch, \emph{i.e.}, over-sample in the low CAS regions while under-sample in the high CAS regions. Hence, the supplementary branch is encouraged to further excavate the action regions underestimated by the base branch, making the localization results more complete. To further promote mutual enhancement and explicitly optimize the localization objective, we construct \textbf{mutual location supervision} between the two branches. Each branch uses location pseudo-labels generated from the CAS of the other branch as localization supervision. We alternately freeze one branch and optimize the other branch, so that the complementary action information of the two branches can be combined to make the location supervision more complete and precise. In multiple iterations, our AMS framework progressively refines the localization results.

In either the base or supplementary branch, we input video features into a backbone network $h(\cdot)$ to predict the category probability and CAS. The backbone network is implemented by a multi-layer perceptron. For the base branch, its backbone network can be formalized as follows:
\begin{equation}
    {\mathbf{M}^{\mathrm{base}}, \, \mathbf{\widehat{y}}^{\mathrm{base}}  = h^{\mathrm{base}}(\mathbf{F}^{\mathrm{orig}}, \, \phi^{\mathrm{base}}),}
\end{equation}
where $\mathbf{M}^{\mathrm{base}}\in\mathbb{R}^{T\times{C}}$ denotes CAS, indicating the probability distribution of each video snippet belonging to all action categories. $\mathbf{\widehat{y}}^{\mathrm{base}}\in\mathbb{R}^{{C}}$ means the predicted video category probability. $\phi^{\mathrm{base}}$ is trainable parameters of the backbone network. In our framework, the input of the base branch is the original feature $\mathbf{F^\mathrm{orig}}$. While the input of the supplementary branch is selected based on the output CAS of the base branch through the adaptive sampler $S(\cdot, \cdot)$; that is,
\begin{equation}
    {\mathbf{F}^{\mathrm{supp}} = S(\mathbf{F}^{\mathrm{orig}}, \, \mathbf{M}^{\mathrm{base}}),}
\end{equation}
where $\mathbf{F}^{\mathrm{supp}} \in\mathbb{R}^{T\times{D}}$. The design details of the sampler are introduced in Section~\ref{subsection: Adaptive Sampling}. Formally, the backbone network of the supplementary branch is formalized as follows:
\begin{equation}
    {\breve{\mathbf{M}}^{{\mathrm{supp}}}, \, \mathbf{\widehat{y}}^{\mathrm{supp}}  = h^{\mathrm{supp}}(\mathbf{F}^{\mathrm{supp}}, \, \phi^{\mathrm{supp}}),}
\end{equation}
where $\mathbf{\widehat{y}}^{\mathrm{supp}}\in\mathbb{R}^{{C}}$ means the predicted category probability, $\phi^{\mathrm{supp}}$ is the trainable parameters, the output CAS $\breve{\mathbf{M}}^{{\mathrm{supp}}}\in\mathbb{R}^{T\times{C}}$. To align the temporal distribution with the CAS of the base branch, we also perform temporal alignment on $\breve{\mathbf{M}}^{{\mathrm{supp}}}$ and obtain ${\mathbf{M}}^{{\mathrm{supp}}}$; see details in Section~\ref{subsection:Temporal Alignment}.


\begin{figure}[t]
\begin{center}
\includegraphics [width=0.48\textwidth] {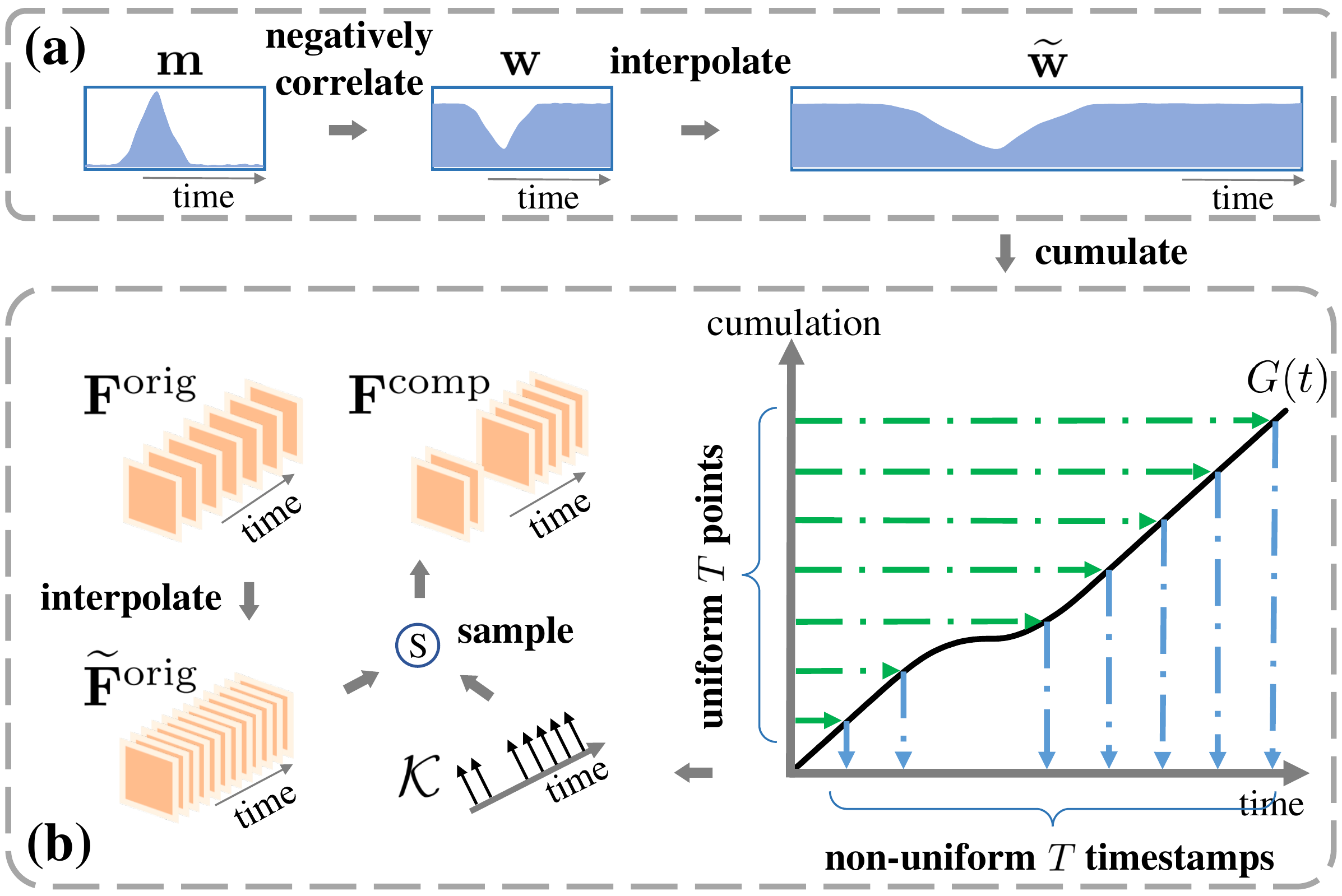}
\end{center}
\vspace{-2pt}
\caption{Illustration of the adaptive sampler.
\textbf{(a): Sampling Weight Sequence.} First calculate the sampling weight sequence $\mathbf{w}$ negatively correlated with the aggregated CAS $\mathbf{m}$, then up-sample $\mathbf{w}$ to get the interpolated weight sequence $\widetilde{\mathbf{w}}$.
\textbf{(b): Sampling Operation.} First cumulate $\widetilde{\mathbf{w}}$ to get its cumulative distribution function $G(t)$; then uniformly sample $T$ points on the cumulation axis, and map them to $G(t)$; finally, map the points on $G(t)$ to the time axis, and obtain the sampling timestamp set $\mathcal{K}$. Sample the interpolated original features $\widetilde{\mathbf{F}}^\mathrm{orig}$ to generate $\mathbf{F}^{\mathrm{supp}}$ for the supplementary branch.}
\label{fig:adaptive sampling}
\vspace{-3pt}
\end{figure}

\subsection{Adaptive Sampler}   \label{subsection: Adaptive Sampling}
\subsubsection{\textbf{Sampling Weight Sequence}}  \label{subsubsection:samplingweightsequence}
Here we propose the implementation details of the adaptive sampler $S(\cdot, \cdot)$. To solve the incompleteness issue of CAS generated by the classifier, we desire to strengthen complementarity between the two branches. To achieve this, we design a novel adaptive sampler to differentiate the inputs of the two branches. Specifically, we input the original video features into the base branch, while dynamically select snippet features for the supplementary branch via the adaptive sampler. The whole process is divided into two parts as shown in Fig.~\ref{fig:adaptive sampling}.

Similar to previous studies~\cite{zeng2019breaking,min2020adversarial,zhong2018step}, the CAS of the base branch tends to focus on the most discriminative action regions. To make the supplementary branch purposefully localize the less discriminative action regions underestimated by the base branch, we need to calculate the class-agnostic sampling weight sequence, based on the CAS of the base branch. Since CAS is a class-specific action probability sequence, to generate the class-agnostic sequence, it is necessary to aggregate all action information of category channels in CAS. Specifically, given the CAS of the base branch $\mathbf{M}^{\mathrm{base}} \in\mathbb{R}^{T\times{C}}$, we only keep the channels of ground truth categories, then perform the maximum operation on these channels in the temporal dimension. We also empirically compare the maximum, average, and random operations in Table~\ref{tab:aggregationmodes}. Formally, the aggregated CAS is denoted as $\mathbf{m} = \{ m_t \} \in\mathbb{R}^{T}$.

According to the meaning of CAS~\cite{shou2018autoloc,paul2018w}, the aggregated CAS $\mathbf{m}$ indicates the action confidences of all snippets in the temporal dimension. In other words, a higher value of $m_t$ indicates a higher confidence of existing an action at the $t$-th snippet. Considering the incompleteness issue of CAS, there might exist some missing actions (less discriminative actions) in the low-value regions of the aggregated CAS. Therefore, we aim to make the supplementary branch focus more on the low-value regions while less on the high-value regions. And the sampling weight sequence is hence designed to be negatively correlated with the aggregated CAS:
\begin{equation}  \label{eq:samplingweight}
    {\mathbf{w} = \{ w_t \} = {\max}(\mathbf{m})- {\mathbf{m}} + \eta \in \mathbb{R}^{{T}},}
\end{equation}
where $\eta$ means a sampling adjustment value, $\max(\cdot)$ is the maximum operations. Each element in the sampling weight sequence $\mathbf{w}$ represents the probability that the corresponding snippet will be selected by the sampling operation. A lower value of the aggregated CAS ${m}_t$ corresponds to a higher probability of ${w}_t$, which indicates that the $t$-th snippet with lower action confidence determined by the base branch is more likely to be sampled. With such a sampling weight sequence, we can naturally over-sample the snippets in the low-value regions of the aggregated CAS while under-sample in the corresponding high-value regions.

\subsubsection{\textbf{Sampling Operation}}
In this section, we perform the sampling operation to generate the inputs for the supplementary branch. Since the original features only contain $T$ snippets, to achieve more fine-grained sampling in the temporal dimension, we first up-sample them by linear interpolation, and then, based on the sampling weight sequence, adaptively select $T$ snippet features from the interpolated original features to form the inputs for the supplementary branch. The interpolated original features are denoted as $\widetilde{\mathbf{F}}^\mathrm{orig}\in\mathbb{R}^{{HT}\times D}$, where $H$ is the interpolation factor. To match the temporal length, we also calculate the interpolated sampling weight sequence, which is denoted as $\widetilde{\mathbf{w}}\in\mathbb{R}^{{HT}}$.

Next, we detail the sampling timestamps and the sampling features in turn, as shown in Fig.\,\ref{fig:adaptive sampling}\,(b).

\noindent \textbf{Sampling Timestamps.} Following the inverse transformation theory~\cite{devroye1986sample}, we adopt a cumulation-mapping manner to adaptively select $T$ snippet features. Concretely, regarding the interpolated weight sequence $\widetilde{\mathbf{w}}$ as a probability mass function, we first cumulate it along the temporal dimension to obtain the cumulative distribution function $G(t)$:
\begin{equation}  \label{eq:cumulativedistributionfunction}
    { G(t) = \sum_{\tau=1}^{t} \widetilde{w}_{\tau} d{\tau} .}
\end{equation}
Intuitively, $G(t)$ corresponds to the black curve in Fig.\,\ref{fig:adaptive sampling}\,(b). Its role is to map the sampling probability of $HT$ snippets uniformly distributed in the temporal dimension, in proportion to the interval length of the cumulation axis. A larger sampling probability $w_t$ indicates a longer interval on the cumulation axis. Hence, sampling $T$ timestamps on the time axis based on the interpolated weight sequence $\widetilde{\mathbf{w}}$ is equivalent to uniformly sample $T$ points on the cumulation axis.

To achieve the sampling goal, we first uniformly sample $T$ points on the cumulation axis; then, map these points to the cumulative distribution function $G(t)$; next, we map the points on $G(t)$ to the time axis; and finally, the corresponding $T$ timestamps on the time axis constitute a candidate sampling timestamp set, denoted as $\mathcal{K}$.

\noindent \textbf{Sampling Features.} Afterward, based on the sampling timestamp set $\mathcal{K}$, we obtain the input features of the supplementary branch from the interpolated original features $\widetilde{\mathbf{F}}^\mathrm{orig}$:
\begin{equation}
    {\mathbf{F}^{\mathrm{supp}} = \mathcal{K} \ {\ltimes} \ \widetilde{\mathbf{F}}^\mathrm{orig}\in \mathbb{R}^{{T\times D}},}
\end{equation}
where $\ltimes$ denotes the index operation, that is, chronologically take out the features corresponding to $T$ sampling timestamps. Accordingly, $\mathbf{F}^{\mathrm{supp}}$ is mainly composed of the features corresponding to the low-CAS value regions of the base branch. Intuitively, the efficacy of adaptive sampler can be interpreted as slowing down the video in the low-CAS regions of the base branch while speeding up the video in the high-CAS regions. The supplementary branch is thus prompted to focus more on less discriminative regions, and further excavates missing actions to complete localization results.

\subsubsection{\textbf{Temporal Alignment}}  \label{subsection:Temporal Alignment}
Since the input features of the supplementary branch are under-sampled or over-sampled in the temporal dimension, they have a non-uniform temporal distribution. Hence, the temporal distribution of the corresponding CAS also becomes non-uniform, which does not match the uniform distribution of the CAS from the base branch. To ensure the localization results of the two branches have the same temporal distribution, we need to make temporal alignment between these two CASs.

For this purpose, we uniformize the temporal distribution for the CAS of the supplementary branch. Concretely, for each timestamp in the uniform distribution, we search the two nearest timestamps on the CAS of the supplementary branch. After that, we can obtain the temporal alignment results by performing linear interpolation between these two CAS values. We denote the temporal alignment operation as $A$, hence the above process is formalized as:
\begin{equation}
    {\mathbf{M}^{\mathrm{supp}} = A(\breve{\mathbf{M}}^{{\mathrm{supp}}})\in\mathbb{R}^{{T\times C}},}
\end{equation}
where $\breve{\mathbf{M}}^{{\mathrm{supp}}}$ is the output CAS of the backbone network from the supplementary branch, $\mathbf{M}^{\mathrm{supp}}$ is its aligned CAS.

\subsubsection{\textbf{Discussion}}
Note that in the proposed adaptive sampler, if the sampling weights of high CAS regions are fixed at 0\%, and the others are 100\%, our sampling strategy will become similar to the existing erase operation~\cite{su2018cascaded,zhong2018step,zeng2019breaking,min2020adversarial}. The proposed adaptive sampler outperforms the erase operation from two aspects. (i) Our sampling probability is soft, while the erase probability is binary. This means that we always retain some action regions determined by the base branch, which act as action anchors for the supplementary branch to avoid paying attention to the background; while the erase operation removes all action regions found previously, usually misleading the model to the background. (ii) Our sampling strategy slows down the less discriminative action regions and speeds up the most discriminative regions, while the erase operation only erases the most discriminative action regions. This means that the inputs generated by our strategy are more fine-grained in the temporal dimension and more purposeful for less discriminative actions, which can lead to better localization results.

\subsection{Mutual Location Supervision}
To promote mutual enhancement and explicitly optimize the localization objective, we further construct mutual location supervision between the base branch and the supplementary branch. Different from the self-training-based strategy~\cite{pardo2021refineloc,zhai2020two,luo2020weakly}, we force the two branches to provide location pseudo-labels for each other, and progressively refine the localization results in multiple iterations.

\subsubsection{\textbf{Location Pseudo-labels}}
After making temporal alignment for the two branches, we encourage them to generate location pseudo-labels from CAS. Since CAS stands for the action confidence probability, a higher value of CAS means a higher confidence of existing an action. To avoid low-quality labels and eliminate uncertainty, we threshold CAS through a hyperparameter $\alpha$ to generate binary pseudo-labels, \emph{i.e.}, if the CAS value of a snippet for any ground truth category is greater than $\alpha$, the snippet is regarded as a positive action example; otherwise, it is considered as a negative action example. The location pseudo-labels can be formulated as:
\begin{equation}
\overline{m}_t^k = \left\{ 
\begin{aligned}
1, & \quad \mathrm{if}  \ {m}_t^k  \; \textgreater \; \alpha \ \mathrm{and} \ y^k=1, \\ 
0, & \quad \mathrm{otherwise}, \\ 
\end{aligned}
\right.
\end{equation}
where ${m}_t^k$ is the CAS value of the $t$-th snippet and the $k$-th category channel, $\overline{m}_t^k$ is the corresponding pseudo-label.

\subsubsection{\textbf{Optimization Process}}
To promote mutual enhancement between these two branches, we construct mutual location supervision, that is, force each branch to leverage the location pseudo-labels from the other branch as the localization objective. The optimization process is achieved by alternately freezing one branch and training the other branch. Specifically, in Phase zero, we train the base branch with only video-level category labels, to generate initial location pseudo-labels. In Phase one, we freeze the base branch, and produce the inputs for the supplementary branch through the adaptive sampler; then optimize the supplementary branch with category labels and location pseudo-labels from the base branch; finally, update pseudo-labels based on the CAS from the supplementary branch. And in Phase two, we optimize the base branch with category labels and the updated location pseudo-labels from the supplementary branch.

To optimize the whole framework, we apply multiple iterations, since one single iteration brings limited improvement to mine less discriminative action regions. In each iteration, the adaptive sampler differentiates the inputs of the two branches, so that they purposefully focus on different action regions. Then, the mutual location supervision obtains more complete location supervision by pushing the CASs of the two branches to be consistent. In the next iteration, the consistent CASs force the adaptive sampler to further update the inputs of the supplementary branch in turn, so that more missing action regions can be explored. Consequently, the adaptive sampler and mutual location supervision jointly contribute to more complete results in progressive iterations.

\subsubsection{\textbf{Loss Function}}
For localization, following~\cite{lin2018bsn,zhao2020bottom,lin2019bmn}, we calculate the weighted cross-entropy loss between location pseudo-labels and the output CAS:
\begin{equation}
    {L_{\mathrm{local}} = \frac{1}{C}\sum_{k=1}^C (\frac{1}{T^{+}} \sum_{t\in \Lambda^{+}} \mathcal{H}({m}_t^k, \overline{{m}}_t^k) +  \frac{1}{T^{-}} \sum_{t\in \Lambda^{-}} \mathcal{H}({m}_t^k, \overline{{m}}_t^k))}
\end{equation}
where $C$ is the number of action categories, ${{m}}_t^k \in [0,1]$ is the output CAS of the $t$-th snippet and the $k$-th category channel, $\overline{m}_t^k \in \{0,1\}$ is the location pseudo-label from the other branch, $\mathcal{H}$ is the regular cross-entropy loss, $\Lambda^{+}$ and $\Lambda^{-}$ denote the positive and negative sample sets, $T^{+}$ and $T^{-}$ are the number of positive and negative samples.

For classification, we calculate the cross-entropy loss between the action category label $\mathbf{y}=[y^1, ..., y^C]^\mathrm{T}$ and the predicted category probability $\widehat{\mathbf{y}} \in \mathbb{R}^{{C}}$:
\begin{equation}
    {L_{\mathrm{class}} = {\frac{1}{C}}\sum_{k=1}^C \mathcal{H}(\widehat{y}^k, y^k),}
\end{equation}
where $\widehat{\mathbf{y}}$ is calculated by aggregating CAS with top-$k$ mean technique~\cite{paul2018w,lee2019background,lee2020background}. For better classification, we combine the classification loss $L_{\mathrm{class}}$ with the Co-Activity Similarity loss in WTALC~\cite{paul2018w}, as the basic loss $L_{\mathrm{basic}}$ of each branch.

Finally, during the training of the whole framework, we combine the basic loss and the localization loss to optimize the base branch or the supplementary branch:
\begin{equation}
 {L_{\mathrm{total}} = L_{\mathrm{basic}} + \lambda L_{\mathrm{local}},}
\end{equation}
where $\lambda$ denotes a balance hyperparameter.

\subsection{Inference}
The AMS framework is separately optimized with RGB and flow features, and the final localization results are generated in a late-fusion fashion. During inference, for an input video, we fuse the two CASs from RGB and flow modes, then average the CASs of the two branches as the final predicted CAS $\mathbf{M}^{\mathrm{final}} \in\mathbb{R}^{T\times{C}}$. The process is given by:
\begin{equation}
 {\mathbf{M}^{\mathrm{final}} = \frac{1}{2}(\mathbf{M}_{\mathrm{flow}}^{\mathrm{base}} + \mathbf{M}_{\mathrm{flow}}^{\mathrm{supp}} + \beta \mathbf{M}_{\mathrm{rgb}}^{\mathrm{base}} + \beta \mathbf{M}_{\mathrm{rgb}}^{\mathrm{supp}}),}
\end{equation}
where $\beta$ is a fusion hyperparameter. After that, we aggregate $\mathbf{M}^{\mathrm{final}}$ to derive the video-level action category probabilities. For classification, we only select the classes whose category probabilities are above the classification threshold $\theta_{cls}$. For the remaining categories, we directly threshold $\mathbf{M}^{\mathrm{final}}$ with the localization threshold $\theta_{loc}$, then concatenate consecutive candidate snippets as action proposals $\{(s_j,e_j,c_j,p_j)\}_{j=1}^o$, where $o$ is the number of proposals, $s_j$, $e_j$, $c_j$, $p_j$ represent the start time, the end time, the action category, and the localization score of the $j$-th action proposal, respectively. The action category $c_j$ of the $j$-th proposal is the action category of the corresponding video. And the localization score $p_j$ of the $j$-th proposal is calculated by the maximum value of $\mathbf{M}^{\mathrm{final}}$ within the proposal interval $[s_j, e_j]$.

\begin{table*}[h]
\small
\caption{Comparison with the state-of-the-art methods on THUMOS14. AVG(0.1-0.5) and AVG(0.3-0.7) are the average mAP from IoU 0.1 to 0.5 and from IoU 0.3 to 0.7, respectively. `Single' denotes training with one single iteration. `Multiple' denotes training with multiple iterations. In general, the performance of the multiple iteration methods is better than that of the single iteration methods. The proposed AMS method outperforms the state-of-the-art methods in the video-level weakly-supervised setting, while performs comparably with several strongly-supervised methods.}
\begin{center}
\begin{tabular}{C{1.7cm}|C{1.4cm}|C{2cm}|C{0.7cm}C{0.7cm}C{0.7cm}C{0.7cm}C{0.7cm}C{0.7cm}C{0.7cm}|C{1cm}|C{1cm}}
\toprule              
\multirow{2}{*}{\textbf{\begin{tabular}[c]{@{}c@{}}Supervision\end{tabular}}} &  \multirow{2}{*}{\textbf{\begin{tabular}[c]{@{}c@{}}Training\end{tabular}}}  &  \multirow{2}{*}{\textbf{Method}} & \multicolumn{7}{c|}{\textbf{mAP@IoU}} & \multirow{2}{*}{\textbf{\begin{tabular}[c]{@{}c@{}}AVG\\ (0.1-0.5)\end{tabular}}} & \multirow{2}{*}{\textbf{\begin{tabular}[c]{@{}c@{}}AVG\\ (0.3-0.7)\end{tabular}}} \\ \cline{4-10}
 &  &  & \textbf{0.1} & \textbf{0.2} & \textbf{0.3} & \textbf{0.4} & \textbf{0.5} & \textbf{0.6} & \textbf{0.7} &  &  \\  \hline \hline
\multirow{6}{*}{\textbf{\begin{tabular}[c]{@{}c@{}}Strong\end{tabular}}}  
 & \multirow{6}{*}{\textbf{\begin{tabular}[c]{@{}c@{}} - \end{tabular}}} & SCNN~\cite{shou2016temporal} & 47.7 & 43.5 & 36.3 & 28.7 & 19.0 & 10.3 & 5.3 & 35.0 & 19.9 \\
 & & TURN~\cite{gao2017turn} & 54.0  & 50.9 & 44.1 & 34.9 & 25.6 & 14.6 & 7.7 & 44.8 & 25.4 \\
 & & SSN~\cite{zhao2017temporal} & \textbf{66.0} & 59.4 & 51.9 & 41.0 & 29.8 & 19.6 & 10.7 & 49.6 & 30.6 \\
 & & A2Net~\cite{yang2020revisiting} & 61.1 & \textbf{60.2} & \textbf{58.6} & \textbf{54.1} & \textbf{45.5} & 32.5 & 17.2 & \textbf{55.9} & 41.6 \\
 & & TAL-Net~\cite{chao2018rethinking} & 59.8 & 57.1 & 53.2 & 48.5 & 42.8 & 33.8 & 20.8 & 52.3 & 41.3 \\ 
 & & BUMR~\cite{zhao2020bottom} & 58.2 & 56.8 & 53.9 & 50.7 & 45.4 & \textbf{38.0} & \textbf{28.5} & 53.0 & \textbf{43.3} \\ 
\hline \hline
\multirow{2}{*}{\textbf{\begin{tabular}[c]{@{}c@{}}Weak\\Count-level\end{tabular}}}
 & \multirow{2}{*}{\textbf{\begin{tabular}[c]{@{}c@{}} - \end{tabular}}} & STARN~\cite{xu2018segregated} & 68.8 & 60.0 & 48.7 & 34.7 & 23.0 & 11.7 & 6.2 & 47.0 & 24.9 \\ 
 & & 3C-Net~\cite{narayan20193c} & 59.1 & 53.5 & 44.2 & 34.1 & 26.6 & 16.7 & 8.1 & 43.5 &25.9 \\
\hline \hline
\multirow{16}{*}{\textbf{\begin{tabular}[c]{@{}c@{}}Weak\\Video-level\end{tabular}}}
 & \multirow{12}{*}{\textbf{\begin{tabular}[c]{@{}c@{}}Single\end{tabular}}} 
 & STPN~\cite{nguyen2018weakly} & 45.3 & 38.8 & 31.1 & 23.5 & 16.2 & 9.8 & 5.1 & 31.0 & 17.1 \\
 & & CPMN~\cite{su2018cascaded} & 47.1 & 41.6 & 32.8 & 24.7 & 16.1 & 10.1 & 5.5 & 32.5 & 17.8 \\
 & & WO~\cite{zeng2019breaking} & 57.6 & 48.9 & 38.9 & 29.3 & 20.5 & - & - & 39.0 & - \\
 & & WTALC~\cite{paul2018w} & 55.2 & 49.6 & 40.1 & 31.1 & 22.8 & 14.8 & 7.6 & 39.8 & 23.3 \\
 & & Autoloc~\cite{shou2018autoloc} & - & - & 35.8 & 29.0 & 21.2 & 13.4 & 5.8 & - & 21.0 \\
 & & Cleannet~\cite{liu2019weakly} & - & - & 37.0 & 30.9 & 23.9 & 13.9 & 7.1 & - & 22.6 \\
 & & BaSNet~\cite{lee2019background} & 58.2 & 52.3 & 44.6 & 36.0 & 27.0 & 18.6 & 10.4 & 43.6 & 27.3 \\
 & & CMCS~\cite{liu2019completeness} & 57.4 & 50.8 & 41.2 & 32.1 & 23.1 & 15.0 & 7.0 & 40.9 & 23.7 \\
 & & BM~\cite{nguyen2019weakly} & 64.2 & \textbf{59.5} & 49.1 & 38.4 & 27.5 & 17.3 & 8.6 & 47.7 & 28.2 \\
 & & ASSG~\cite{zhang2019adversarial} & \textbf{65.6} & 59.4 & \textbf{49.5} & 38.7 & 25.4 & 15.0 & 6.6 & 47.7 & 27.0 \\
 & & A2CL-PT~\cite{min2020adversarial}  & 61.2 & 56.1 & 48.1 & \textbf{39.0} & \textbf{30.1} & 19.2 & 10.6 & \textbf{50.7} & \textbf{29.4} \\
 & & DGAM~\cite{shi2020weakly}  & 60.0 & 54.2 & 44.8 & 38.2 & 28.8 & \textbf{19.8} & \textbf{11.4} & 45.2 & 28.6  \\ \cline{2-12}  \cline{2-12} 
 & \multirow{4}{*}{\textbf{\begin{tabular}[c]{@{}c@{}}Multiple\end{tabular}}} & RefineLoc~\cite{pardo2021refineloc}  & - &- & 33.9 & - & 22.1 & - & 6.1 & - & - \\
 & & TSCN~\cite{zhai2020two}  & 63.4 & 57.6 & 47.8 & 37.7 & 28.7 & 19.4 & 10.2 & 47.0 & 28.8    \\
 & & EM-ML~\cite{luo2020weakly} & 59.1 & 52.7 & 45.5 & 36.8 & 30.5 & 22.7 & \textbf{16.4} & 44.9 & 30.4 \\
\cline{3-12} 
 & & \textbf{AMS (Ours)} & \textbf{69.1} & \textbf{62.3} & \textbf{52.7} & \textbf{42.8} & \textbf{33.1} & \textbf{23.1} & 13.0 & \textbf{52.0} & \textbf{32.4} \\
\bottomrule  
\end{tabular}
\label{tab:results on THUMOS14}
\end{center}
\end{table*}


\begin{table}[t] 
\small
\centering 
\caption{Comparison with the state-of-the-art methods on ActivityNet1.2. AVG(0.5-0.95) denotes the average mAP at IoU thresholds 0.5:0.05:0.95. `Single' means training with one single iteration. `Multiple' means training with multiple iterations. The proposed AMS method outperforms all previous methods in terms of the average mAP, while surpasses most methods at some IoU thresholds.}
\begin{tabular}{C{1.8cm}|C{1.9cm}|C{0.4cm}C{0.4cm}C{0.4cm}|C{1.2cm}}
\toprule
\multirow{2}{*}{\textbf{\begin{tabular}[c]{@{}c@{}}Supervision\\(Training)\end{tabular}}} &  \multirow{2}{*}{\textbf{Method}} & \multicolumn{3}{c|}{\textbf{mAP@IoU}} & \multirow{2}{*}{\textbf{\begin{tabular}[c]{@{}c@{}}AVG\\ (0.5-0.95)\end{tabular}}} \\ \cline{3-5}
  & & \textbf{0.5} & \textbf{0.75} & \textbf{0.95} 
\\ \hline \hline
\multirow{2}{*}{\textbf{\begin{tabular}[c]{@{}c@{}} Strong \end{tabular}}} 
& CDC~\cite{shou2017cdc}  & \textbf{45.3}  & 26.0 & 0.2   & \textbf{23.8}\\ 
& SSN~\cite{zhao2017temporal}  & 41.3  & \textbf{27.0} & \textbf{6.1}   & \textbf{26.6}\\ 
\hline \hline
\multirow{8}{*}{\textbf{\begin{tabular}[c]{@{}c@{}} Weak\\Video-level\\(Single) \end{tabular}}}
 & U-Nets~\cite{wang2017untrimmednets} & 7.4   & 3.2    & 0.7   & 3.6 \\
 & Autoloc~\cite{shou2018autoloc}      & 27.3  & 15.1   & 3.3   & 16.0 \\
 & TSM~\cite{yu2019temporal}           & 28.3  & 17.0   & 3.5   & 17.1    \\
 & WTALC~\cite{paul2018w}              & 37.0  & 12.7   & 4.5   & 18.0    \\
 & Cleannet~\cite{liu2019weakly}       & 37.1  & 20.3   & 5.0   & 21.6 \\
 & CMCS~\cite{liu2019completeness}     & 36.8  & 22.0   & \textbf{5.6}  & 22.4 \\
 & BaSNet~\cite{lee2019background} & 38.5  &\textbf{24.2} &\textbf{5.6} &24.3 \\
 & DGAM~\cite{shi2020weakly}  & \textbf{41.0}  &23.5 &5.3 &\textbf{24.4}  \\
\hline
\multirow{4}{*}{\textbf{\begin{tabular}[c]{@{}c@{}} Weak\\Video-level\\(Multiple) \end{tabular}}}
 & EM-ML~\cite{luo2020weakly} & 37.4   &-  &- & 20.3 \\ 
 & RefineLoc~\cite{pardo2021refineloc}  & 38.0   &20.8  &4.9 & 22.2 \\ 
 & TSCN~\cite{zhai2020two}  & 37.6  &\textbf{23.7} & 5.7 & 23.6 \\
 \cline{2-6} 
 & \textbf{AMS (Ours)}   &\textbf{40.7}   &\textbf{23.7}   & \textbf{5.8}  & \textbf{24.6} \\ 
\bottomrule 
\end{tabular}
\label{tab:results on Activitynet}
\end{table}


\section{Experimental Results}
\label{section:experiments}
In this section, we conduct extensive experiments to evaluate the effectiveness of our AMS method and reveal the effect of each component. We first detail the experimental settings and network architectures, then report the corresponding experimental results. Based on two widely used datasets, \emph{i.e.}, THUMOS14~\cite{jiang2014thumos} and ActivityNet~\cite{caba2015activitynet}, our AMS method significantly outperforms previous state-of-the-art methods both quantitatively and qualitatively. Besides that, both adaptive sampler and mutual location supervision strategy have a great effect on the localization performance.

\subsection{Dataset and Evaluation}
\noindent \textbf{THUMOS14~\cite{jiang2014thumos}.}
There are 413 untrimmed videos in 20 categories, and each video contains an average of 15 action instances. The model is trained on 200 validation videos, and evaluated on 213 test videos. This dataset is widely used and challenging, because the video lengths vary widely and the actions occur very frequently.

\noindent \textbf{ActivityNet1.2~\cite{caba2015activitynet}.}
The dataset contains 9682 videos belonging to 100 categories, which are divided into 4619 videos for training, 2383 videos for validation, and 2480 videos for testing. Since the ground-truth action intervals of test videos are not available, we train the model on the training set and evaluate on the validation set. Almost all videos contain only a single action category, and action regions take up more than half of the duration in most videos.

\noindent \textbf{Evaluation Metrics.}
Following the convention, we evaluate our method with the standard mean Average Precision (mAP) at different thresholds of temporal intersection over union (T-IoU). Note that a proposal is regarded as positive only if both the predicted category is correct and T-IoU exceeds the set threshold. Besides, each ground-truth action instance can only match one action proposal. The mAP is calculated from the evaluation code provided by the corresponding datasets.


\begin{table*}[t]
\centering
\caption{Contribution of three components, \emph{i.e.}, the branch number, the adaptive sampler, and location supervision on THUMOS14. `Single' means training with one single iteration. `Multiple' means training with multiple iterations. AVG(0.1-0.7) denotes the average mAP from IoU 0.1 to 0.7. There are two ways to provide location supervision: mutual location supervision and self-training-based location supervision~\cite{pardo2021refineloc,zhai2020two}, which are abbreviated as `mutual' and `self', respectively. Both the adaptive sampler and mutual location supervision have great effects on the localization performance.}
\begin{tabular}{C{1.5cm}|C{0.5cm}|C{0.8cm}C{0.8cm}C{1.5cm}|C{0.6cm}C{0.6cm}C{0.6cm}C{0.6cm}C{0.6cm}C{0.6cm}C{0.6cm}|C{1.2cm}}
\toprule
 &  &  & &  & \multicolumn{7}{c|}{mAP@IoU} &  \\ \cline{6-12}
\multirow{-2}{*}{Training} &  \multirow{-2}{*}{ID} & \multirow{-2}{*}{\textbf{\begin{tabular}[c]{@{}c@{}}Branch\\ number\end{tabular}}} & \multirow{-2}{*}{\textbf{\begin{tabular}[c]{@{}c@{}}Adaptive\\ sampler\end{tabular}}} & \multirow{-2}{*}{\textbf{\begin{tabular}[c]{@{}c@{}}Location\\ supervision\end{tabular}}} & 0.1 & 0.2 & 0.3 & 0.4 & 0.5 & 0.6 & 0.7 & \multirow{-2}{*}{\begin{tabular}[c]{@{}c@{}}AVG\\ (0.1-0.7)\end{tabular}} \\ \hline
\multirow{3}{*}{\begin{tabular}[c]{@{}l@{}}Single\end{tabular}} 
& (A) & one  & no & no & 62.1 & 53.7 & 43.1 & 33.3 & 24.7 & 15.6 & 8.1 & 34.4 \\  
& (B) & dual & no & no & 62.4 & 54.1 & 43.5 & 33.4 & 24.9 & 15.8 & 8.1 & 34.6 \\ 
& (C) & dual & yes & no & 67.3 & 58.2 & 47.0 & 36.3 & 27.3 & 17.8 & 9.5 & 37.7 \\ 
\hline
\multirow{3}{*}{\begin{tabular}[c]{@{}l@{}}Multiple\end{tabular}} 
& (D) & dual & no & self & 65.2 & 57.4 & 47.1 & 37.4 & 29.4 & 19.3 & 10.6 & 38.0 \\
& (E) & dual & no & mutual & 65.8 & 58.1 & 47.7 & 37.9 & 29.8 & 19.7 & 10.9 & 38.6 \\ 
& (F) & dual & yes & mutual & \textbf{69.1} & \textbf{62.3} & \textbf{52.7} & \textbf{42.8} & \textbf{33.1} & \textbf{23.1} & \textbf{13.0} & \textbf{42.3} \\ 
\bottomrule
\end{tabular}
\label{tab:Contribution of three components}
\end{table*}


\subsection{Implementation Details}
\noindent \textbf{Feature Extraction.} 
Following previous methods~\cite{zeng2019breaking,su2018cascaded,liu2019completeness}, to reduce the computational requirements, we extract the high-level features of the input video in advance, and then train the whole framework with these high-level features. Due to the memory constraint, we first split the input video into non-overlapping $16$-frame snippets, then randomly sample $T$ consecutive snippets from each video, since the video lengths vary greatly. $T$ is set to $1000$ on THUMOS14, and $400$ on ActivityNet1.2. We leverage the TV-L1 algorithm~\cite{wedel2009improved} to extract optical flow from RGB data; next, we use the two-stream I3D architecture~\cite{carreira2017quo} pre-trained on the Kinetics dataset~\cite{carreira2017quo} to extract RGB and flow features; and finally, we can obtain the 1024-dimensional feature from RGB data or flow data of each video snippet.

\noindent \textbf{Backbone Network.}
In either the base branch or the supplementary branch, the backbone network maps video features to CAS and video category probabilities. Structurally, it cascades a feature transformation module and a mapping module. The former consists of a fully connected layer, followed by ReLU activation and Dropout to fine-tune the video features from the feature extractor. The latter contains two parallel fully connected layers, followed by the softmax function respectively, to predict video category probabilities and CAS.

\noindent \textbf{Implementation.}
The proposed framework is implemented with Pytorch~\cite{paszke2017automatic}, using Adam optimizer~\cite{kingma2014adam} with the learning rate of $10^{-4}$ to respectively optimize on THUMOS14 and ActivityNet1.2. For a fair comparison, we fix the pre-trained parameters of the feature extractor without fine-tuning. We train the framework for 20 epochs in Phase zero, then alternately train the two branches every 5 epochs in Phase one and Phase two. All hyperparameters are determined by grid search: the balance hyperparameter $\lambda$ = 1.0, the fusion hyperparameter $\beta$ = 0.15, the sampling adjustment value $\eta$ = 0.75, the interpolation factor $H$ = 20, the classification threshold $\theta_{cls}$ = 0.25. The threshold $\alpha$ for generating location pseudo-labels is set equal to the localization threshold $\theta_{loc}$, which is calculated adaptively by $0.7\times{\mathrm{avg}(\mathbf{M})}$.

\subsection{Comparison with State-of-the-Art Methods}
Under multiple IoU thresholds, we compare the proposed AMS method with existing methods from various levels of supervision settings. As introduced in Section~\ref{section:introduction}, existing weakly-supervised temporal action localization methods can be divided into two categories: classification-based framework and self-training-based framework. The main difference is that the latter requires multiple iterations for training compared to the former. We thus abbreviate these two frameworks as the `single' and `multiple' groups. Since our AMS method also uses pseudo-labels for multiple iterations, we put it into the `multiple' group for a fair comparison.

Table~\ref{tab:results on THUMOS14} reports the comparison on THUMOS14. In terms of performance, the multiple iteration methods are generally better than the single iteration methods. And our AMS method achieves a new state-of-the-art in the video-level weakly-supervised setting. Compared to previous multiple iteration methods, \emph{i.e.},~\cite{luo2020weakly, pardo2021refineloc, zhai2020two}, our method obtains a performance gain of more than 4.9\% in terms of the average mAP from IoU 0.1 to 0.5, while more than 2.0\% in terms of the average mAP from IoU 0.3 to 0.7. This proves the effectiveness of our proposed framework, and means that our AMS method produces more precise and complete localization. Moreover, despite being trained in the weakly supervised setting, our method performs comparably with several early strongly-supervised methods~\cite{shou2016temporal,gao2017turn,zhao2017temporal}.

Table~\ref{tab:results on Activitynet} also reports the results on ActivityNet1.2. Generally speaking, our AMS method surpasses most of previous methods under the same level of supervision. In terms of the average mAP, our method outperforms all existing weakly-supervised methods, and follows strongly-supervised SSN~\cite{zhao2017temporal} with the least gap. Note that, compared with THUMOS14, ActivityNet has only one-tenth of action instances per video on average, and almost all videos contain only one action category. Therefore, this dataset has a lower localization requirement, just as emphasized in~\cite{chao2018rethinking,zhang2019adversarial,luo2020weakly}. This lower requirement might lead to the small gain of our method to some extent. All in all, the great performance on the two datasets indicates the effectiveness of our method.

\subsection{Ablation Studies}
Here we conduct ablation studies on THUMOS14 to analyze the effect of each component in our framework, \emph{i.e.}, the branch number, mutual location supervision, and the adaptive sampler. Recently, the self-training-based strategy~\cite{pardo2021refineloc,luo2020weakly,zhai2020two} has been proposed to explicitly optimize the localization objective. It adopts the pseudo-labels generated by CAS in the current step as the location supervision for the next step, and relies on the self-training strategy to iteratively refine the localization results. For a detailed comparison, we also reproduce this method under the same settings.

Specifically, we experiment with the following six setups. The baseline is set as a vanilla classification-based method~\cite{nguyen2018weakly,paul2018w}, which is trained with only the basic loss $L_\mathrm{basic}$, then thresholds CAS for localization results. 
(A): The baseline with only one branch;
(B): The baseline with two identical branches;
(C): Add the adaptive sampler on (B);
(D): Add the self-training-based strategy on (B);
(E): Add the mutual location supervision strategy on (B);
(F): Add the adaptive sampler and the mutual location supervision strategy on (B).
For the above setups, as the two branches could generate quite different CASs, we average their CASs as the final output CAS. Table~\ref{tab:Contribution of three components} reports all the localization results.

\subsubsection{One branch v.s. Dual branch}
In the proposed AMS framework, the dual-branch setting causes the model parameters to be doubled. We thus explore the effect of the branch number on the localization performance. Comparing (B) to (A), we can find that there is no significant difference between the localization performance of the dual-branch model and the one-branch model. The only 0.2\% average mAP improvement suggests that simply doubling the model parameters cannot bring a significant increase in performance. Besides, (A) and (B) perform worst among these six setups. This is because relying only on classification supervision can mislead the model to focus on the most discriminative regions, resulting in sparse and incomplete localization results.

\subsubsection{Self-training-based strategy v.s. Mutual location supervision}
To evaluate the effect of location pseudo-labels, we add either mutual location supervision or self-training-based strategy to (B), and perform multiple training iterations. Comparing (D) or (E) to (B), we see that multiple iterations with location supervision improve the performance by 3.4\% average mAP. This reflects that explicitly optimizing the localization objective can reconcile the contradiction between classification and localization in the WTAL task.

Besides, comparing (E) to (D), our proposed mutual location supervision outperforms the self-training-based strategy by 0.5\% average mAP. We conjecture the reason as follows. During multiple iterations, the self-training-based strategy separately trains the two branches. While in each iteration, our mutual location supervision promotes mutual enhancement between the two branches, and combines the action information from the two branches, which plays the role of ensemble learning~\cite{polikar2012ensemble,zhou2009ensemble,dietterich2002ensemble} to bring a certain improvement.

\subsubsection{Effectiveness of the adaptive sampler}
We also add the adaptive sampler on (B) and (E), to verify the effectiveness of the proposed sampling strategy. Comparing (C) to (B), the adaptive sampler purposefully differentiates the inputs of the two branches, and brings a gain of 3.1\% average mAP. By adaptively selecting less discriminative snippets for the supplementary branch, the sampler promotes it to explore the actions underestimated by the base branch, thus completing the localization results of the whole framework.

Moreover, comparing (F) to (E), the adaptive sampler further boosts the effectiveness of mutual location supervision in multiple iterations, bringing a gain of 3.7\% average mAP. By integrating the adaptive sampler and mutual location supervision to form our AMS framework, we achieve the best performance with large gaps from the others, which indicates that both components play essential roles and jointly contribute to more complete action localization results.

\subsection{Validation and Analysis Experiments}
\subsubsection{Effect of the adaptive sampling weights}
As designed in Section~\ref{subsubsection:samplingweightsequence}, we design the sampling weight sequence to be negatively correlated with the CAS of the base branch. To verify the effectiveness of this strategy, we implement three experiments in Table~\ref{tab:samplingweight}. (A): uniformly generate sample weights; (B): randomly generate sample weights; (C): adaptively generate sample weights by our proposed strategy. The results show that the adaptive sampling strategy significantly surpasses the other two, boosting the performance to more than 3.4\% average mAP. As is evident, the adaptive sampling strategy effectively differentiates the two branches, prompting the supplementary branch to complement the detection results of the base branch. Moreover, (B) outperforms (A) by 0.8\% average mAP. We speculate this is because random sampling acts as a role of data augmentation, which enlarges the amount of training data, thus bringing a certain gain.

\begin{table}[t]
\caption{Results of different sampling weight strategies on THUMOS14. AVG(0.1-0.7) means the average mAP from IoU 0.1 to 0.7. Our proposed adaptive sampling is significantly better than random sampling and uniform sampling.}
\begin{center}
\begin{tabular}{C{0.6cm}|C{1.2cm}|C{0.5cm}C{0.5cm}C{0.5cm}C{0.5cm}|C{1.1cm}}
\toprule
\multirow{2}{*}{ID} & \multirow{2}{*}{\begin{tabular}[c]{@{}c@{}}Sampling\\weights\end{tabular}} & \multicolumn{4}{c|}{mAP@IoU} & \multirow{2}{*}{\begin{tabular}[c]{@{}c@{}}AVG\\ (0.1-0.7)\end{tabular}} \\ \cline{3-6}
 & & 0.1 & 0.3 & 0.5 & 0.7 &  \\ \hline \hline
(A) & {Uniform} & 65.3 & 47.0 & 29.3 & 10.7 & 38.1  \\
(B) & {Random} & 66.7 & 47.8 & 29.8 & 11.1  & 38.9  \\ 
(C) & {Adaptive} & \textbf{69.1} & \textbf{52.7} & \textbf{33.1} & \textbf{13.0} & \textbf{42.3}\\
\bottomrule
\end{tabular}
\end{center}
\label{tab:samplingweight}
\end{table}


\begin{figure}[t]
\centering
\vspace{7pt}
\centerline{\includegraphics[width=0.4\textwidth]{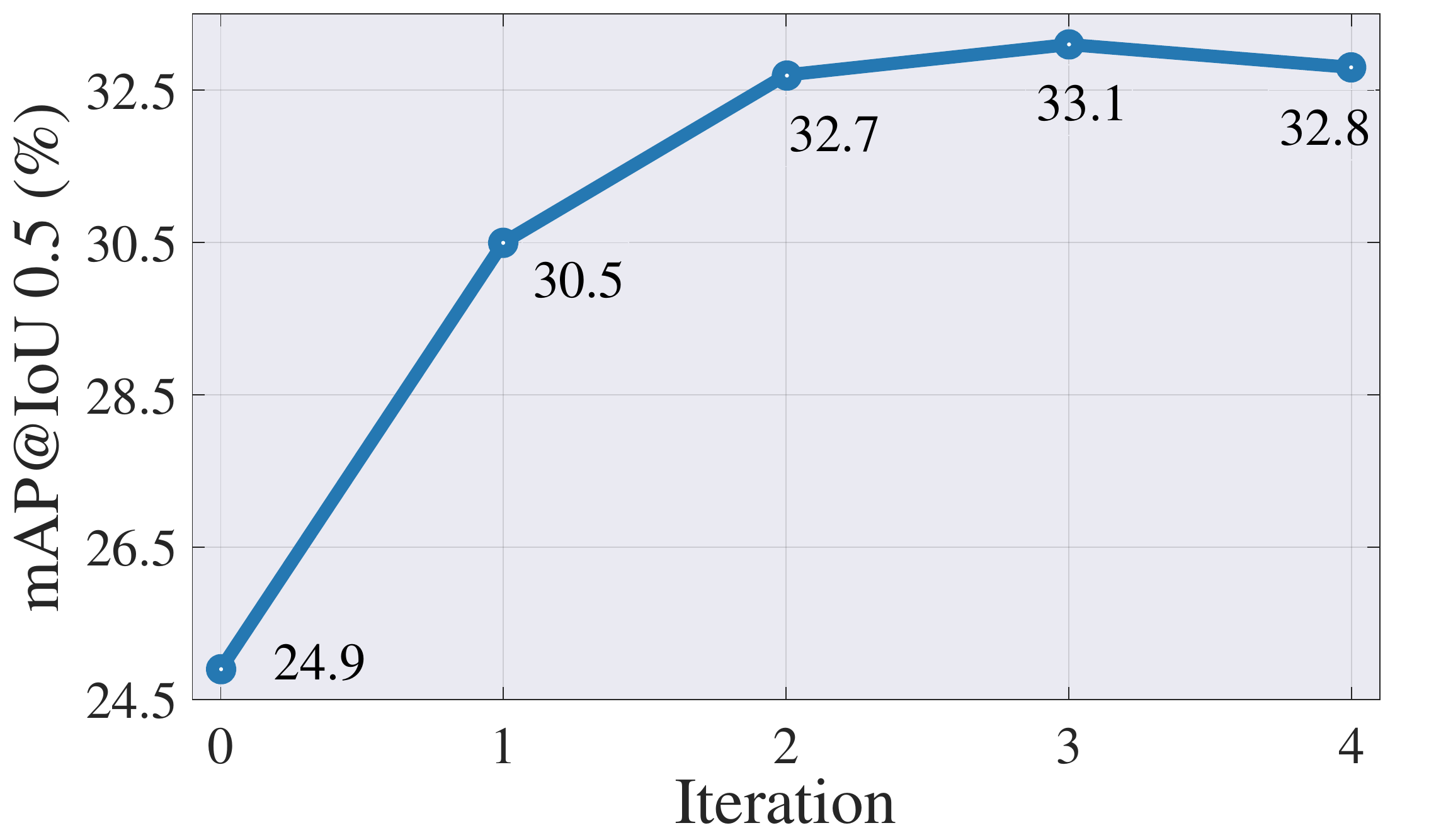}} 
\caption{Results of progressive iterative refinement on THUMOS14. The best performance is obtained in the third iteration, and the improvement over four iterations is 8.2\% mAP.}
\label{fig:iterate}
\end{figure}

\begin{table}[t]
\caption{Results of three aggregation operations in Section~\ref{subsubsection:samplingweightsequence} on THUMOS14. `Random' means randomly selecting a ground-truth category channel. Both the `Maximum' operation and the `Average' operation bring promising results.}
\begin{center}
\begin{tabular}{C{1.5cm}|C{0.6cm}C{0.6cm}C{0.6cm}C{0.6cm}|C{1.1cm}}
\toprule
\multirow{2}{*}{\begin{tabular}[c]{@{}c@{}}aggregation\\method\end{tabular}} & \multicolumn{4}{c|}{mAP@IoU} & \multirow{2}{*}{\begin{tabular}[c]{@{}c@{}}AVG\\ (0.1-0.7)\end{tabular}} \\ \cline{2-5}
 & 0.1 & 0.3 & 0.5 & 0.7 &  \\ \hline \hline
{Random} & 67.5 & 51.6 & 32.5 & 12.5 & 41.4  \\
{Average} & 68.9 & 52.5 & \textbf{33.3} & \textbf{13.1}  & 42.2  \\ 
{Maximum} & \textbf{69.1} & \textbf{52.7} & 33.1 & 13.0 & \textbf{42.3} \\
\bottomrule
\end{tabular}
\end{center}
\label{tab:aggregationmodes}
\end{table}

\begin{figure}[t]
\centering
\vspace{6pt}
\centerline{\includegraphics[width=0.39\textwidth]{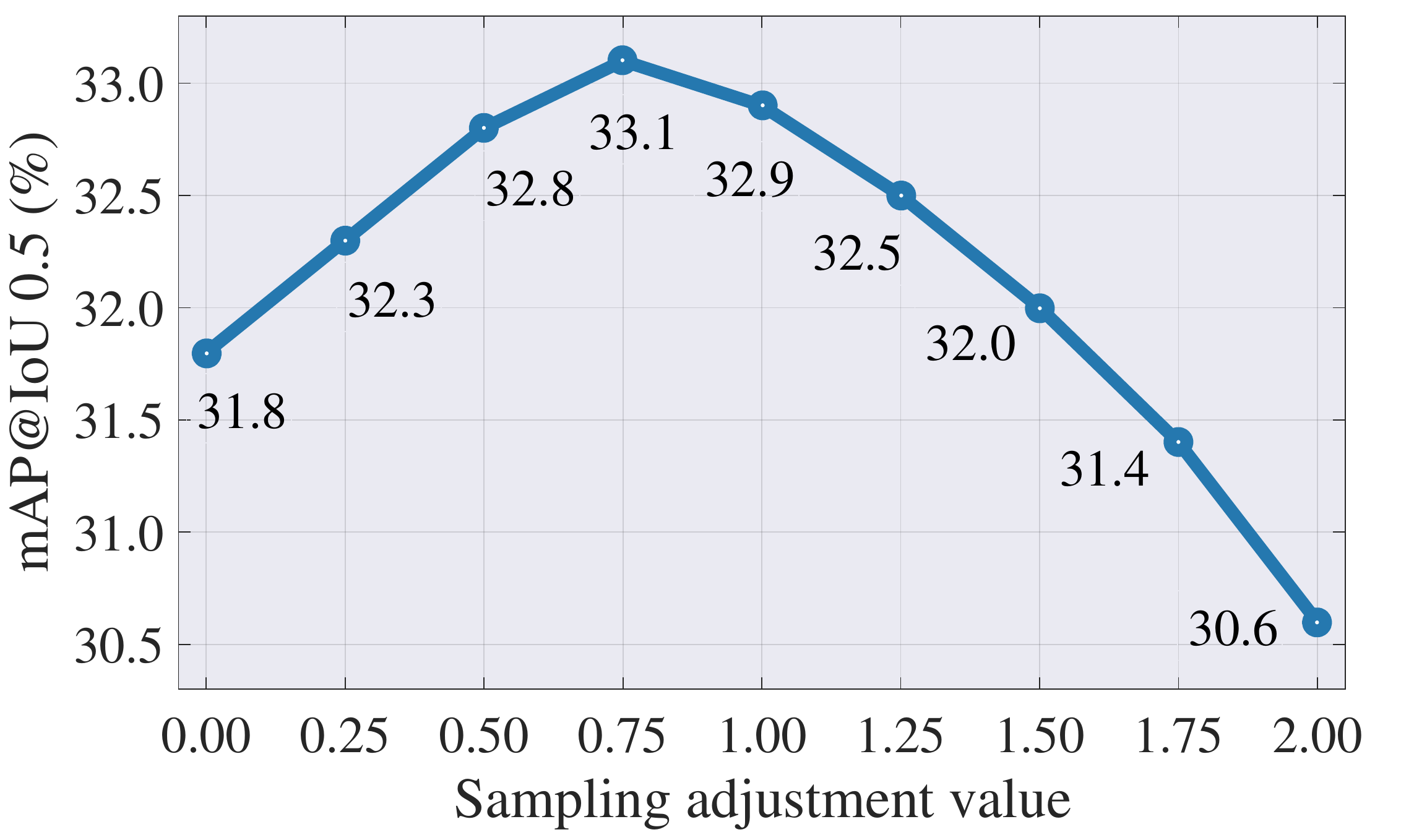}} 
\caption{Effects of the sampling adjustment value $\eta$ on THUMOS14. The localization results are reported at IoU threshold 0.5. And the best performance is achieved when $\eta=0.75$.}
\label{fig:constant}
\end{figure}


\subsubsection{Effect of the progressive iterative refinement}
Fig.~\ref{fig:iterate} quantifies the results of four iterations to evaluate the effect of progressive refinement. The best performance is obtained in the third iteration, which gains 8.2\% mAP compared to the initial iteration. Such a huge gain shows that in progressive iterations, the quality of pseudo-labels is continuously improving through mutual enhancement between the two branches. And the adaptive sampler and mutual location supervision can collaborate and promote each other. In essence, the mutual enhancement is a voting ensemble of the localization results from the two branches, which can provide more complete and precise supervision, compared to each individual branch. When we combine the location information of the two branches, the localization errors that only exist in one branch are largely ruled out, thus avoiding error propagation. 

Moreover, we notice that the gain of a single iteration is decreasing until it becomes zero. In the initial iteration, the location pseudo-labels are low-quality, or even none. While after three iterations, the quality of pseudo-labels tends to be high and stable, the performance reaches the main bound.

\subsubsection{Effect of the aggregation operations}
As described in Section~\ref{subsubsection:samplingweightsequence}, we aggregate all action information of category channels in CAS through the average operation. There are two other aggregation operations available, that is, the maximum operation and the operation randomly selecting a ground-truth category channel. Table~\ref{tab:aggregationmodes} summarizes the comparison. There is no significant performance difference between the average operation and the maximum operation. And the results of these two operations are superior to the random operation. The reason may be that the random operation also provides the supplementary branch with some action regions that have been found by the base branch. This increases the overlap between the actions discovered by the supplementary branch and the base branch, thus damaging the performance.

\begin{figure*}[t]
\centering
\centerline{\includegraphics[width=1.0\textwidth]{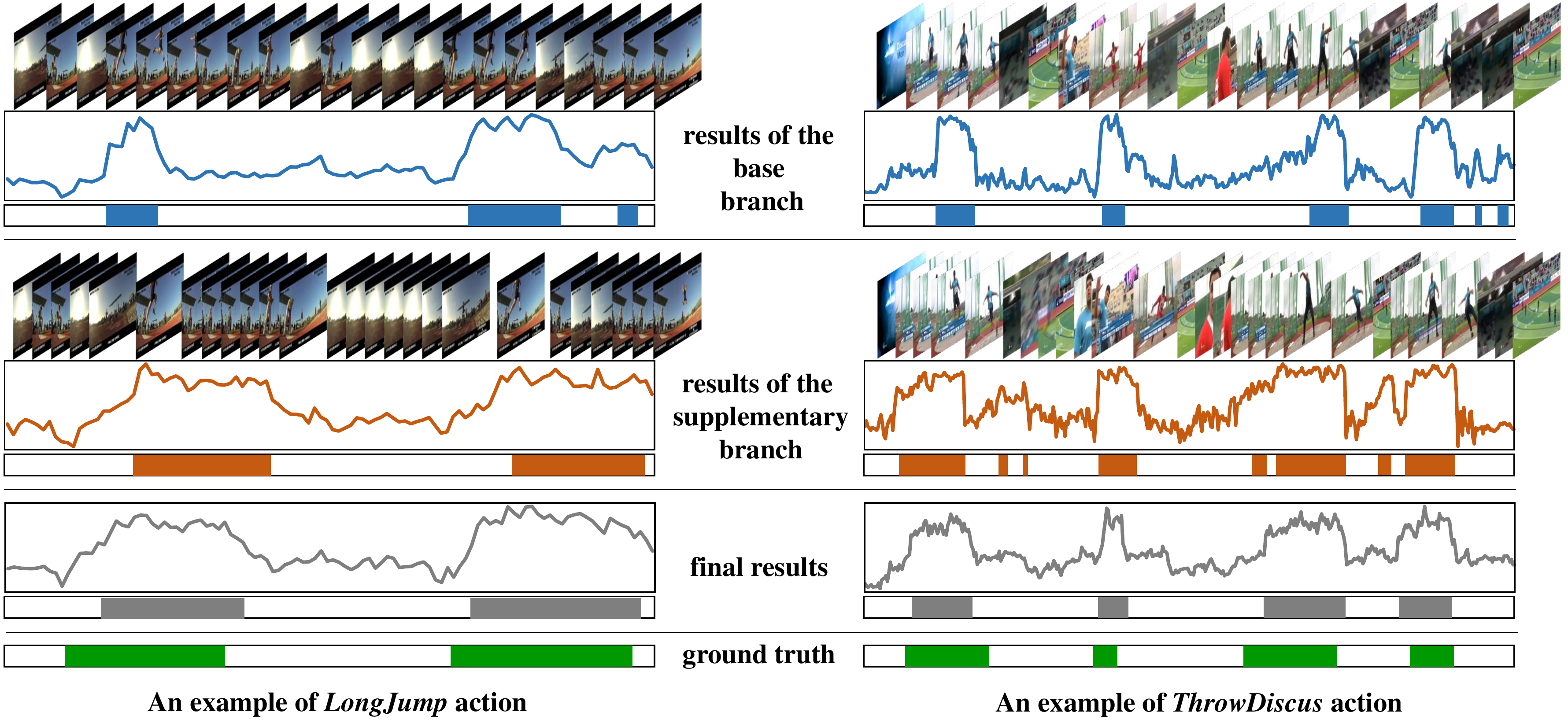}} 
\vspace{5pt}
\caption{Qualitative results on THUMOS14 (Best viewed in color). In each example, there are nine plots. The first three plots are the input video, the CAS, and localization results of the base branch. The middle three plots show the input video, the CAS, and localization results of the supplementary branch. The last three plots are the final CAS, the final localization results of the whole framework, and the ground truth action intervals. The base branch, fed with videos of uniformly temporal distribution, can only detect the most discriminative actions. Through the adaptive sampler, we input videos of non-uniformly temporal distribution to the supplementary branch, hence force it to purposefully complement the less discriminative actions underestimated by the base branch. Mutual location supervision makes our final localization results more complete and precise.}
\vspace{5pt}
\label{fig: Qualitative Results}
\end{figure*}

\subsubsection{Effect of the sampling adjustment value $\eta$}
In Eq.~\ref{eq:samplingweight}, the sampling adjustment value $\eta$ is designed to adjust the sampling weight sequence. Fig.~\ref{fig:constant} demonstrates the effect, where $\eta$ varies from 0 to 2 with an interval of 0.25. We find that it shows a clear trend to peak at 0.75. And there are slight differences in performance when using values from 0.5 to 1. However, when we continue to increase $\eta$ to 2, the performance drops a lot. This is because in this case, the sampling weights of different video snippets tend to be the same, causing the adaptive sampling to degenerate to the uniform sampling. Accordingly, the inputs of the two branches become substantially identical, and our AMS framework degenerates into the straightforward mutual location supervision model.

\subsection{Qualitative Results}
To qualitatively demonstrate the superiority of the proposed framework, we visualize several examples in Fig.~\ref{fig: Qualitative Results}. For clear understanding, we provide the inputs, the CAS, and the localization results of the base branch and the supplementary branch in turn. Generally speaking, whether for videos containing sparse or dense action instances, the localization results of our framework are relatively complete and precise. More specifically, the base branch, which is fed with videos of uniformly temporal distribution, can only detect the most discriminative action regions. Therefore, its corresponding results are sparse and trivial. Relying on the adaptive sampler, we select the uncertain video snippets of the base branch as inputs for the supplementary branch. The videos of non-uniformly temporal distribution, force the supplementary branch to purposefully complement the less discriminative actions underestimated by the base branch, but also causes some false-positive background predictions. On this basis, mutual location supervision promotes the mutual enhancement between the two branches, which combines their localization results for more complete and precise final prediction results. The good qualitative results again prove the effectiveness of our proposed framework.

\section{Conclusion}  \label{section:conclusion}
In this work, to solve the incompleteness issue of CAS in WTAL, we propose an adaptive mutual supervision framework (AMS) with two branches. The base branch leverages CAS to localize the most discriminative action regions, and the supplementary branch localizes the less discriminative action regions through a novel adaptive sampler. The adaptive sampler dynamically updates the input of the supplementary branch with a sampling weight sequence negatively correlated with the CAS from the base branch, thus prompting the supplementary branch to localize the action regions underestimated by the base branch. To promote mutual enhancement between the two branches, we construct mutual location supervision. Each branch uses location pseudo-labels generated from the other branch as localization supervision. By alternately optimizing the two branches in multiple iterations, we progressively localize more complete action regions. Experiments on two benchmarks demonstrated the effectiveness and outstanding performance of the proposed AMS framework.

\ifCLASSOPTIONcaptionsoff
  \newpage
\fi

\bibliographystyle{IEEEtran}
\bibliography{refer}

\end{document}